\documentclass[10pt,twocolumn,letterpaper]{article}

\usepackage{wacv}
\usepackage{times}
\usepackage{epsfig}
\usepackage{graphicx}
\usepackage{amsmath}
\usepackage{amssymb}

\usepackage{caption}
\usepackage{hhline}
\usepackage{booktabs}
\usepackage{pifont}
\usepackage{adjustbox}
\newcommand{\xmark}{\ding{55}}

  {\begin{list}{$\bullet$}%
     {\topsep=0in\itemsep=0pt\parsep=0pt\partopsep=0in}%
   }{\end{list}\addvspace{0pt}}

\newcommand{\ignore}[1]{}

\usepackage[pagebackref=true,breaklinks=true,letterpaper=true,colorlinks,bookmarks=false]{hyperref}

\wacvfinalcopy 

\def\wacvPaperID{386} 

\begin{document}

\title{Photo-Sketching:\\ Inferring Contour Drawings from Images}
\author{Mengtian Li\textsuperscript{1} \qquad Zhe Lin\textsuperscript{2} \qquad Radom\'ir M\v ech\textsuperscript{2} \qquad Ersin Yumer\textsuperscript{3}  \qquad Deva Ramanan\textsuperscript{1,4}\\
\textsuperscript{1}Carnegie Mellon University \qquad \textsuperscript{2}Adobe Research  \qquad \textsuperscript{3}Uber ATG  \qquad \textsuperscript{4}Argo AI
}


\twocolumn[{
\renewcommand\twocolumn[1][]{#1}
\maketitle
\begin{center}
    \centering
    \includegraphics[width=1\textwidth]{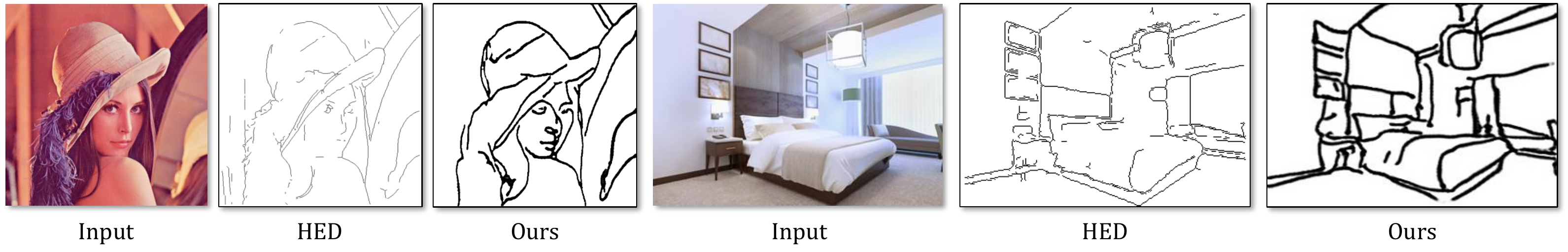}
   \captionof{figure}{Automatic contour drawing generation for images in the wild. Different from traditional edge or boundary detectors \cite{xie15hed}, our method predicts the most salient contours in images and reflects imperfections in ground truth human drawing, \eg the ceiling is not perfectly straight in the right example as a novice drawer would draw it. Right photo by ostap25 -- \url{stock.adobe.com}.}
    \label{fig:teaser}
    \bigskip
    \bigskip
\end{center}
}]


\begin{abstract}

Edges, boundaries and contours are important subjects of study in both computer graphics and computer vision. On one hand, they are the 2D elements that convey 3D shapes, on the other hand, they are indicative of occlusion events and thus separation of objects or semantic concepts. In this paper, we aim to generate contour drawings, boundary-like drawings that capture the outline of the visual scene. Prior art often cast this problem as boundary detection. However, the set of visual cues presented in the boundary detection output are different from the ones in contour drawings, and also the artistic style is ignored. We address these issues by collecting a new dataset of contour drawings and proposing a learning-based method that resolves diversity in the annotation and, unlike boundary detectors, can work with imperfect alignment of the annotation and the actual ground truth. Our method surpasses previous methods quantitatively and qualitatively. Surprisingly, when our model fine-tunes on BSDS500, we achieve the state-of-the-art performance in salient boundary detection, suggesting contour drawing might be a scalable alternative to boundary annotation, which at the same time is easier and more interesting for annotators to draw.


\end{abstract}

\section{Introduction}


\begin{figure*}[t]
\begin{center}
\includegraphics[width=1\linewidth]{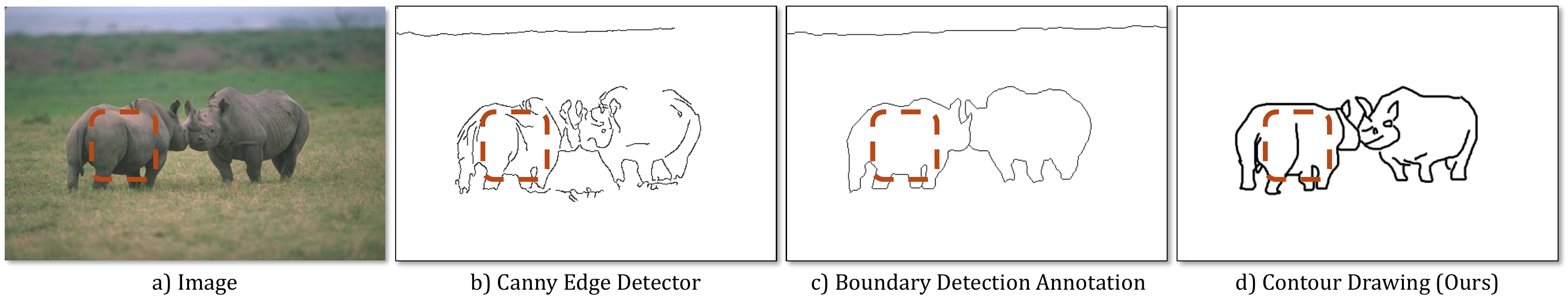}
\end{center}
\vspace{-1em}
\caption{a) Which visual cues would you draw when sketching out an image? b) Traditional edge detectors \cite{Canny1986ACA} only capture high frequency signals in the image without image understanding. c) Boundary detectors are usually trained on edges that derived from closed segment annotations and therefore, they do not include salient inner boundaries by definition \cite{Martin2001ADO,amfm_pami2011}. d) In contrast, our contour drawing (ground truth is shown here) contains both the occluding contours and salient inner edges. For example, the dashed box in the top row contains a open contour ending in a {\em cusp}~\cite{koenderink1990solid,Forsyth2012ComputerV}.}
\label{fig:compareboundary}
\end{figure*}

Edge-like visual representation, appearing in form of image edges, object boundaries, line drawings and pictorial scripts, is of great research interest in both computer vision and computer graphics. Automatic generation of such representation enables us to understand the geometry of the scene \cite{revaud2015epicflow}, and perform image manipulation in this sparse space \cite{Dekel2018SparseS}.
This paper studies such representation in the form of {\em contour drawing}, which contains object boundaries, salient inner edges such as occluding contours, and salient background edges. These sets of visual cues convey 3D perspective, length and width as well as thickness and depth \cite{Sutherland1997Gesture}.
Contour drawings are usually based on real-world objects (immediately observed or from memory), and therefore, can be considered as an expression of human vision. Its counterpart in machine vision is edge and boundary detection. Interesting, the set of visual cues is different in contour drawings and in image boundaries (Fig \ref{fig:compareboundary}). Comparing to image boundaries, contour drawings tend to have more details inside each object (including occluding contours and semantically-salient features such as eyes, mouths, etc.) and are made of strokes that are loosely aligned to pixels on the image edges.
We propose a contour generation algorithm to output contour drawings given input images. This generation process involves identifying salient boundaries and is connected with the salient boundary detection in computer vision. In fact, we will show that our contour generation algorithm can be re-purposed to perform salient boundary detection and achieve the best performance on standard benchmark. Another element involved contour drawing generation is to adopt proper artistic style. Fig \ref{fig:teaser} shows our method successfully captures the style and itself is a style transfer application. Moreover, contour drawing is an intermediate representation between image boundary and abstract line drawing. Our study of contour drawing paves the road for machine's understanding and generation of abstract line drawings \cite{Eitz2012HowDH,Cole2008WhereDP}.

What types of edge-like visual representation are studied in existing work? In non-photorealistic rendering, 2D lines that convey 3D shapes are widely studied. The most important of such might be the {\em occluding contours}, regions where the local surface normal is perpendicular to the viewing direction, and {\em creases}, edges along which the dihedral angle is small \cite{koenderink1990solid}. It is noted by DeCarlo \etal \cite{decarlo2003suggestive} that important details are missing if only those edges are rendered, and their solution is to add {\em suggestive contours}, regions where occluding contour would appear with minimal change in viewpoint. As a result of having clear mathematical definition, these edge-like representation can be directly computed using methods in differential geometry given the 3D model. In computer vision, a different set of visual cues are defined and are inferred from the image alone without knowledge of the 3D world, namely the {\em image edges} and {\em boundaries}. Image edges correspond to sharp changes in image intensity due to changes in albedo, surface orientation, or illumination~\cite{Forsyth2012ComputerV}. Boundaries, as formally defined by Martin \etal \cite{1273918}, are contours in the image plane that represents a change in pixel ownership from one object or surface to another. Nonetheless, this definition ignores the fact that the contour can also appear on a smooth surface of the same object, for example the {\em cusp} in Fig \ref{fig:compareboundary}. Since much progress has been driven by datasets, in practise, the boundaries are ``defined'' by the seminal benchmark of BSDS300 \cite{1273918} and BSDS500 \cite{amfm_pami2011}. Interestingly, despite their popularity, these datasets were originally designed and annotated to be a segmentation dataset. This means that boundaries are derived from closed segments annotated by humans \cite{Martin2001ADO}, and yet not all boundaries form closed shapes.

\begin{figure*}[t!]
\begin{center}
\includegraphics[width=1\linewidth]{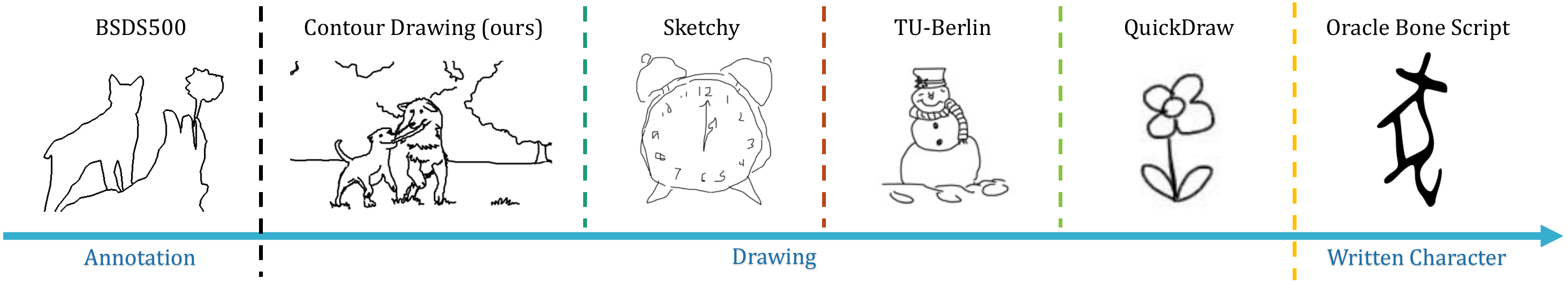}
\end{center}
\vspace{-1em}
\caption{Comparison with other edge-like representations. Here we order examples \cite{amfm_pami2011,sketchy2016,eitz2012hdhso,Ha2018ANR}  by their level of abstraction. Our contour drawing cover more detailed internal boundaries than a boundary annotation, while having much better alignment to actual image contours, and much more complexity compared to other drawing-based representations.}
\vspace{+1em}
\label{fig:sketchcompare}
\end{figure*}

The other related line of research revolves around the representation of sketch. Most work study the relationship between the strokes themselves
without a reference object or image \cite{Ha2018ANR,Eitz2012HowDH,schneider2014sketch}. While some work on sketch-based image retrieval \cite{sketchy2016,eitz2011sketch} and sketch generation \cite{Ha2018ANR,Song2018LearningTS} indeed models the correspondence between sketch and image, the type of drawings used are far too simple or abstract, and does not contain edge-level correspondence, making it unsuitable for training a generic scene sketch generator. A comparison is summarized in Fig \ref{fig:sketchcompare} and Tab \ref{tab:comparesketch}.  

\begin{table*}[t!]
\centering
\adjustbox{width=1\linewidth}{
\begin{tiny} 
\begin{tabular}{l|c|c|c|c|c} \hline
Dataset & Edge Aligned & Multiple Obj & With Image & Vec Graphics & Stroke Order \\ \hline
BSDS500 \cite{amfm_pami2011} & \checkmark & \checkmark & \checkmark & \xmark & \xmark \\ \hline
Contour Drawing (ours) & Roughly & \checkmark & \checkmark & \checkmark & \checkmark \\ \hline
Sketchy \cite{sketchy2016} & \xmark & \xmark & \checkmark & \checkmark & \checkmark \\ \hline
TU-Berlin \cite{eitz2012hdhso} & \xmark & \xmark & \xmark & \checkmark & \checkmark \\ \hline
QuickDraw \cite{Ha2018ANR}  & \xmark & \xmark & \xmark & \checkmark & \checkmark \\ \hline
\end{tabular}
\end{tiny}
}
\caption{Dataset comparison. Our proposed contour drawing dataset is different from prior work in terms of boundary alignment, multiple objects, corresponding image-sketch pairs, vector graphics encoding, and stroke ordering annotation.}
\label{tab:comparesketch}
\end{table*}

To accommodate our research on contour drawings, we collect a dataset containing 5000 drawings (Sec \ref{sec:dataset}). The challenge for training a contour generator is to resolve the diversity among the contours for the same image obtained from multiple annotators. We address it by proposing a novel loss that allows the network to converge to an implicit consensus, while retaining details (Sec \ref{sec:gensketch}). Our contour generator can be applied to salient boundary detection. By simply fine-tuning on BSDS500, we achieve the state-of-the-art performance (Sec \ref{sec:boundarydetection}). Finally, we show our dataset can be expanded in a cost free way with a sketch game (Sec \ref{sec:expansion}). Our code and dataset are available online \footnote{\url{http://www.cs.cmu.edu/~mengtial/proj/sketch}}.

\section{Collecting Contour Sketches} \label{sec:dataset}

We create our novel task with the the popular crowd-sourcing platform Amazon Mechanical Turk \cite{Buhrmester2011AmazonsMT}.
To collect drawings that are roughly boundary aligned, we allow the Turkers to trace over a fainted background image.
In order to obtain high-quality drawings, we design a labeling interface with a detailed instruction page including many positive and negative examples. The quality control is realized through manual inspection by treating drawings of the following types as rejection candidates: (1) missing inner boundary, (2) missing important objects, (3) with large misalignment with original edges, (4) the content not recognizable, (5) drawing humans with stick figures, (6) shaded on empty areas.

Finally, we collect 5000 high-quality drawings on a dataset of 1000 outdoor images crawled from Adobe Stock \cite{adobestock} and each image is paired with exactly 5 drawings. In addition, we have 1947 rejected submissions, which will be used in setting up an automatic quality guard as discussed in Sec \ref{sec:expansion}.




\section{Sketch Generation} \label{sec:gensketch}

In this section, we propose a new deep learning-based model to generate contour sketches from a given image and evaluate it against competing methods in both objective and subjective manner. A unique aspect of our problem here is that each training image is associated with multiple ground truth sketches drawn by different annotators.

\subsection{Previous Methods}


Early methods of line-drawing generation focus on human faces, where they build explicit models to represent facial features \cite{brennan1984caricature,937657}. Other work focuses on generating the style but leaving the task of deciding which edge to draw to the user \cite{kang2005interactive,Xie2015StrokeBasedSL}. More recently, Song \etal \cite{Ha2018ANR} used LSTM to sequentially generate the stroke for simple doodles of several strokes. However, our contour drawings on average contain 44 strokes and around 5,000 control points, way beyond the capacity of existing sequential models.

\subsection{Our Method}

\begin{figure*}[t]
\begin{center}
\includegraphics[width=1\linewidth]{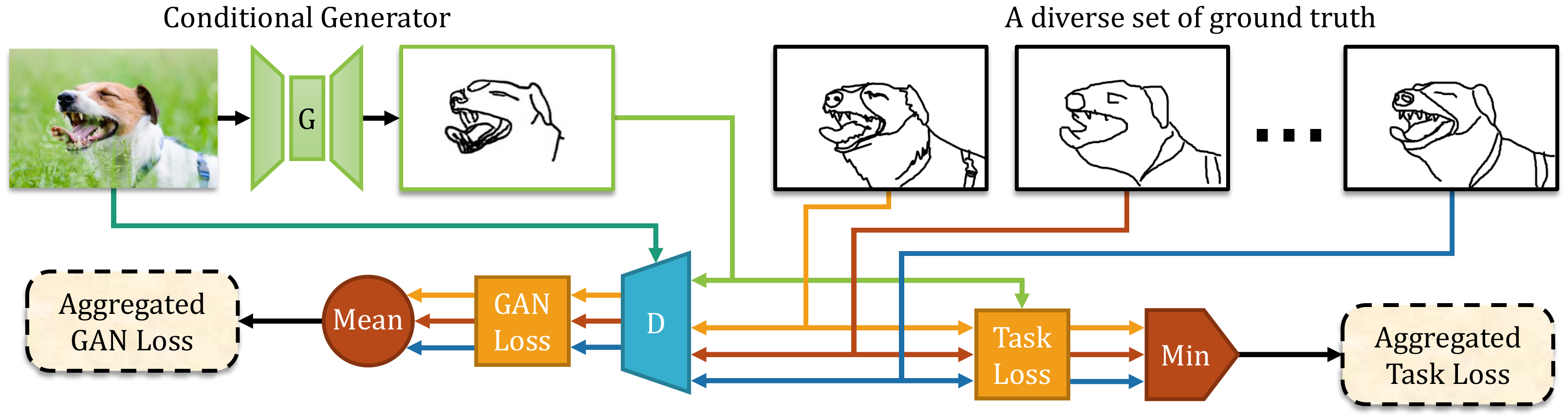}
\end{center}
\caption{We train an image-conditioned contour generator with a novel MM-loss (Min-Mean-loss) that accounts for multiple diverse outputs encountered during training (top row). Training directly on the entire set of image-contour pairs generates conflicting gradients. To rectify this, we carefully aggregate the discriminator ``GAN" loss and the regression ``Task" loss. The discriminator averages the GAN loss across all image-contour pairs, while the regression-loss finds the minimum-cost contour to pair with this image (determined on-the-fly during learning). This ensures that the generator will not simply regress to the ``mean" contour, which might be invalid. 
Photo by alexei\_tm -- \url{stock.adobe.com}.
}
\label{fig:model}
\end{figure*}

Naturally, the problem of generating contour drawing can be cast into an image translation problem or a classical boundary detection problem. Given the popularity of using conditional Generative Adversarial Networks (cGANs) to generate images from sketches or boundary maps, one might think the apparently easier inverse problem can be solved by reversing the image generation direction. However, none of the existing cGAN methods \cite{isola2017image,CycleGAN2017,liu2017unsupervised,zhu2017multimodal} have shown results on such a task and our experiments show that they do not work on sketch generation out-of-the-box. We conjecture that the drawings are sparse and discrete representation compared to textured images. It might be easier to obtain gradients in the latter case. Also, our dataset has more than one target for each source image (1-to-many mapping). And modeling such diversity makes it difficult to optimize. On the other hand, classical boundary detection approaches linearly combines the different ground truths to form a single target per each input. This form of data augmentation bypasses the need to model the diverse outputs and results in a soft output as well, but it is not the case with multiple ground truth having edges not perfectly aligned. The soft representation no longer bears the meaning of boundary strength, but how well the edges are accidentally matched. Training on such data yields unreasonable output for both our method and existing boundary detection methods. Hence, our problem cannot be trivially solved by training or fine-tuning boundary detectors on the contour drawing dataset. Another issue with soft representation, as found in \cite{Hou2013BoundaryDB}, is their poor correlation with the actual boundary strength. We share the same findings in our experiments, as it is difficult to find a single threshold for the final output that works well for all images.
In this work, we use a different cGAN with a novel MM-loss (Fig \ref{fig:model}).

\subsubsection{Formulation}

We leverage the power of the recently popular framework of adversarial training. In a Generative Adversarial Network (GAN), a random noise vector $z$ is fed into the generation network $G$ to generate an output image $y$. In the conditional setup (cGAN), the generator takes input an image $x$, and together with a $z$, it maps to a $y$. The generator $G$ aims to generate ``real" images conditioned on $x$, while there is another discriminator network $D$ that is adversarially trained to tell the generated images from the actual ground truth target. Mathematically, the loss for such objective can be written as

\begin{align}
    \mathcal{L}_{cGAN}(x, y, z) = \min_G\max_D \mathbb{E}_{x,y}[\log D(x,y)] + \notag \\
    \mathbb{E}_{x,z}[\log (1-D(x,G(x,z))]
    \label{eq:cGAN}
\end{align}

As found by previous work \cite{Mathieu2016DeepMV,isola2017image}, the noise vector $z$ is usually ignored in the optimization. Therefore, we do not include $z$ in the our experiments. We also followed the common approach in cGAN to include a task loss in addition to the GAN loss. This is reasonable since we have a target ground truth for us to compare with directly. For our contour generation task, we set the task loss to be L1 loss which encourages sparsity required for contour outputs. The combined loss function now becomes

\begin{align}
    \mathcal{L}_c(x, y) = \lambda \mathcal{L}_{cGAN}(x, y) +  \mathcal{L}_{\texttt{1}}(x, y),
    \label{eq:Lc}
\end{align}
where the the non-negative constant $\lambda$ adjusts the relative strength of the two objectives. Note that when $\lambda = 0$, the model reduces to a simple regression.

The above formulation assumes a 1-to-1 mapping between the two domains. However, we have multiple different targets $y_i^{(1)}, y_i^{(2)}, ..., y_i^{(M_i)}$ for a same input $x_i$, making it a 1-to-many mapped problem. Note the number of targets $M_i$ for each input may vary from examples to examples. If we ignore the fact of 1-to-many mapping, this is reduced to a regular 1-to-1 mapping problem: $(x_1, y_1^{(1)})$, ..., $(x_1, y_1^{(M_1)})$, ..., $(x_N, y_N^{(1)})$, ..., $(x_N, y_N^{(M_i)})$, and those pairs are fetched in random order to train the network.

Our method treats $(x_i, y_i^{(1)}, ..., y_i^{(M_i)})$ as a single training example. To accommodate the extra targets in each training example, we propose a novel MM-loss (Min-Mean-loss) (Fig \ref{fig:model}). Two different aggregate functions are used for the generator $G$ and the discriminator $D$ respectively. The final loss for each training example becomes

\begin{align}
    \mathcal{L}(x_i, y_i^{(1)}, ..., y_i^{(M_i)}) = \frac{\lambda}{M_i} \sum_{j=1}^{M_i}\mathcal{L}_{cGAN}(x_i, y_i^{(j)}) + \notag \\
    \min_{j \in \{1, ..., M_i\}} \mathcal{L}_{\texttt{1}}(x_i, y_i^{(j)}),
    \label{eq:L}
\end{align}

The ``mean'' aggregate function asks the discriminator to learn from all modalities in the target domain and treat those modalities with equal importance. The ``min'' aggregate function allows the generator to adaptively pick the most suitable modality to generate on-the-fly. Therefore, the problem of conflicting gradients caused by different modalities is greatly alleviated. In the diagnostic experiments (Tab \ref{tab:ablation}), we find that training on the consensus drawing outperforms the baseline method, while training on the complete set of sketches with MM-loss outperforms training just on consensus. The ``min'' aggregation function might be reminiscent of the stochastic multiple-choice loss \cite{lee2016stochastic} which relies on a single target output but learns multiple network output branches to generate a diverse output. In our setting, we have a single stochastic output, but multiple ground-truth targets, and a part of the network (the discriminator) that still uses the set of all ground truths to back-propagate the gradients.

\begin{figure}[t]
\begin{center}
\includegraphics[width=1\linewidth]{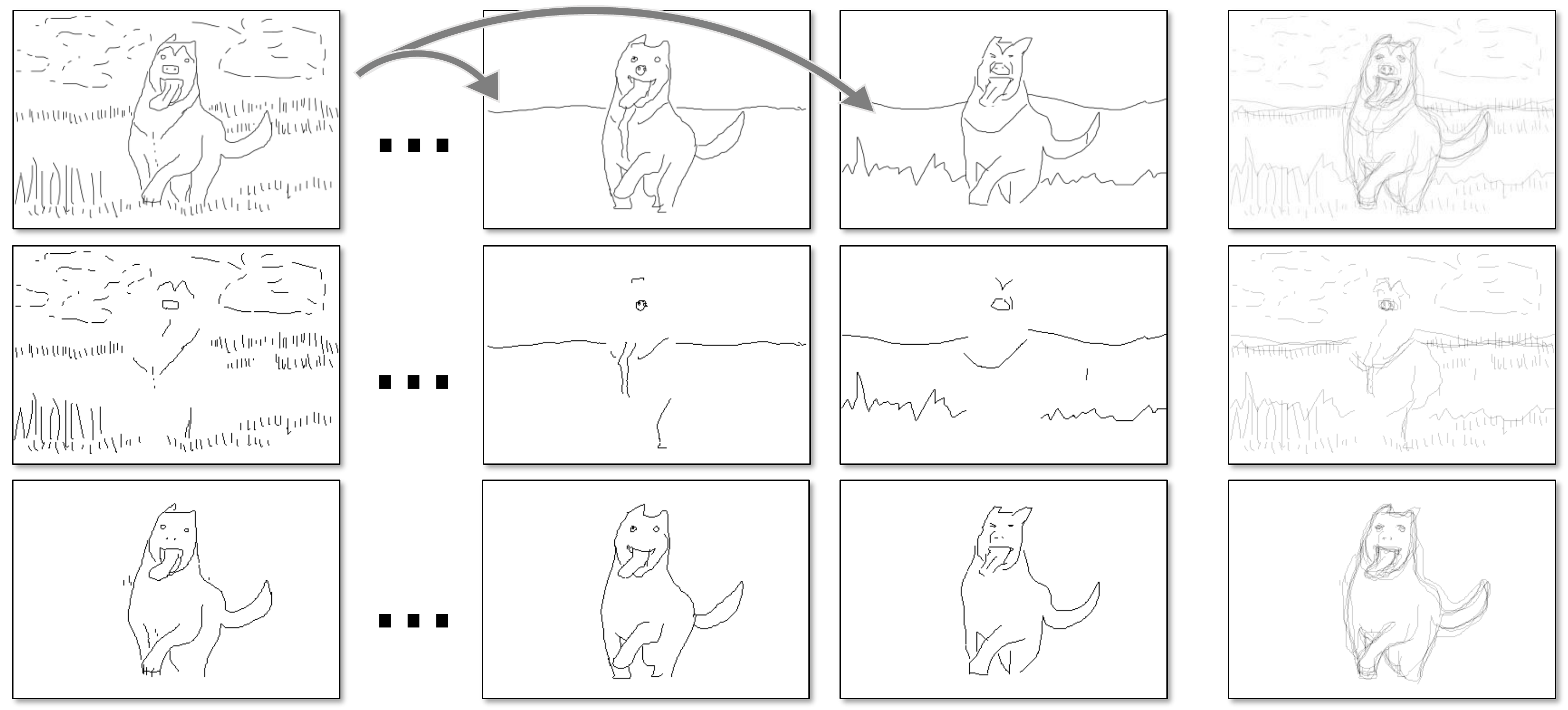}
\end{center}
\vspace{-1em}
\caption{Finding consensus from diverse drawings. In row 1, we visualize 3 different ground truth drawings corresponding to the same image, followed by their overlay in the fourth column. We match strokes in one drawing to another, removing those strokes that could not be matched (row 2). The leftover matched strokes (row 3) are used for evaluation. Note our novel loss allows us to train on the original drawing (row 1) directly and it outperforms the results training on the consensus (row 3), as shown in Tab \ref{tab:ablation} second last row.}
\label{fig:consensus}
\end{figure}

We use the standard encoder-decoder architecture (ResNet \cite{He2016DeepRL} based) that yields good performance on style translation tasks \cite{Johnson2016PerceptualLF}. Unlike other pixel generation tasks, we find the skip connections between the encoder and the decoder make the performance drop. The reason might be that our targets contain mainly object boundaries instead of edges in textures, and removing the skip connections suppresses this low-level information. In many pixel-level prediction tasks, the skip connections are added to make pixel accurate predictions. However, we find that pixel accuracy is already encoded in the network itself since our output is sparse. This can be evidenced by the pixel accurate predictions of our same model applied to boundary detection (Sec \ref{sec:boundarydetection}). For the discriminator, we used a regular global GAN as opposed to PatchGAN \cite{pix2pix2017} in related work. Although PatchGAN helps other networks to generate nice textures, it discourages the network to ``think'' globally, resulting in many broken edges for a single countour of the object. This problem is alleviated when using the global GAN. An ablation study is provided in Tab \ref{tab:ablation} with evaluation metric explained in the next subsection.

\begin{table}[t]
\centering
\begin{tabular}{lrrr}
\toprule
Method     & F1-score & Precision & Recall \\
\midrule
pix2pix \cite{pix2pix2017} (baseline) & 0.514 & 0.585 & 0.458 \\
\midrule
+ ResNet generator & 0.561 & 0.620 & 0.512 \\
+ our MM-loss & 0.765 & 0.814 & 0.722 \\
+ GlobalGAN & 0.773 & 0.720 & 0.835 \\
+ augmentation (\textbf{final}) & {\bf 0.826} & 0.861 & 0.794 \\
\midrule
- train on consensus & 0.802 & 0.915 & 0.714 \\
- remove GAN loss & 0.778 & 0.889 & 0.692 \\
\bottomrule
\end{tabular}
\caption{Ablation study of our method on the validation set. The metrics are explained in Sec \ref{sec:sketchobjeval}. We built up our model from a baseline method \cite{pix2pix2017} and the final model uses the ResNet generator without skip connection, a global discriminator and our proposed MM-loss. Moreover, despite the inconsistency in the non-consensus strokes, training on the original drawings outperforms training on just consensus strokes (second last row). We conjecture that our MM-loss can resolve conflicting supervision on the fly. Also, the last row shows that by adopting adversarial training, we outperform pure regression.}
\label{tab:ablation}
\end{table}

\begin{figure*}[t]
\begin{center}
\includegraphics[width=1\linewidth]{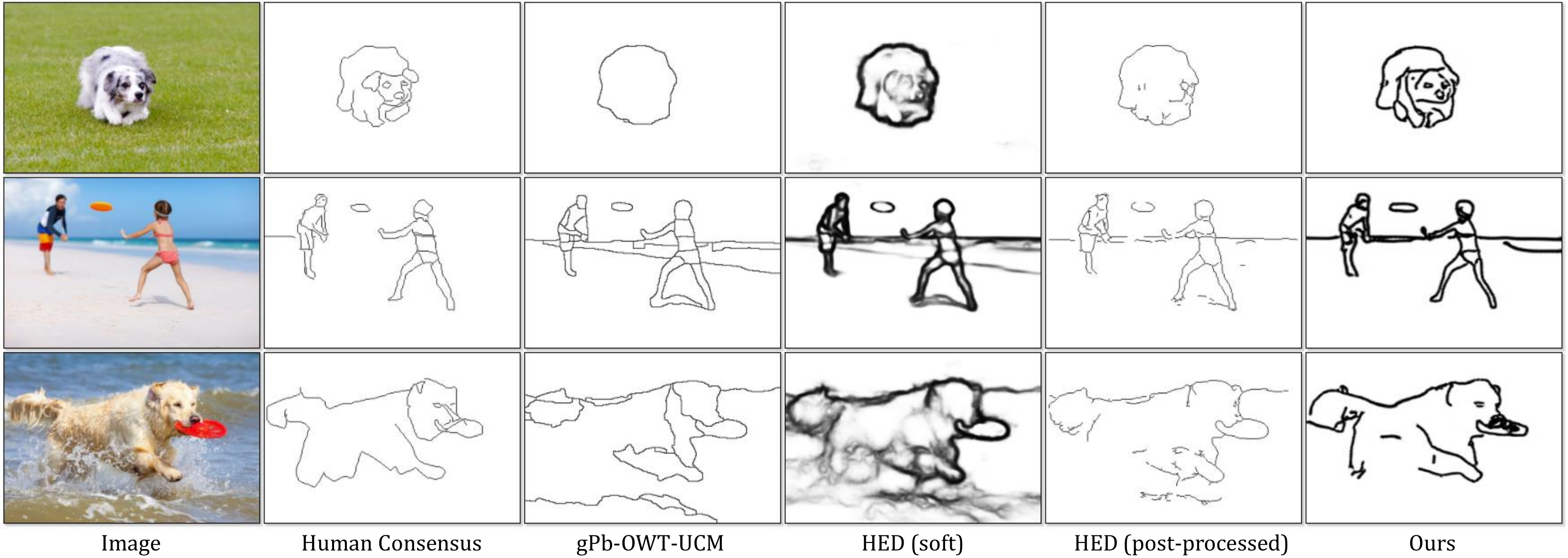}
\end{center}
\vspace{-0.5em}
\caption{Qualitative results for contour drawing generation. Column 3 and 5 are results at their optimal thresholds for the entire test set (\ie ODS, for readers familiar with BSDS evaluation \cite{amfm_pami2011}). Note that it is non-trivial to convert HED \cite{xie15hed} soft output (column 4) to clean sketches (column 5) without artifacts (broken edges). The last row shows that our methods fails at places with partial occlusion (water splash). In contrast, all human annotators are not confused by such occlusion. Photos from top to bottom by martincp, BlueOrange Studio, and Phil Stev -- \url{stock.adobe.com}.}
\label{fig:visualsketch}
\end{figure*}

\subsection{Evaluation}

{\bf Quantitative Evaluation} \label{sec:sketchobjeval}
Boundary detection has a well-established evaluation protocol that matches predicted pixels to the ground truth under a given offset tolerance \cite{1273918,amfm_pami2011}. Matching is done with min-cost bipartite assignment \cite{Goldberg1995,Cherkassky1997}. To apply this approach to contour generation, we first need to reconcile the diverse drawing styles in the ground truth set. Hou \etal \cite{Hou2013BoundaryDB} propose a {\em consensus matching} evaluation of boundary detection that refines the ground-truth by matching pixels from one human annotation to another, removing those that are not unanimously matched across all annotators. We follow suit, but match at stroke level to ensure that strokes are not broken up in the final consensus drawing (Fig \ref{fig:consensus}). In addition, since contour drawings are not exactly aligned with the image boundary, we double the standard offset tolerance used for boundary evaluation. The evaluation treats each ground truth pixel as an ``object" in the precision-recall framework. We split the set of 1000 images with associated sketches into train-val-test sets of 800-100-100. The results are shown in Tab~\ref{tab:sketch} and Fig \ref{fig:visualsketch}. Pix2Pix and our method are trained on our dataset while HED is off-the-shelf. As explained earlier, boundary detection methods cannot work with diverse ground truths and imperfect alignment between the annotations and fine-tuning them on our dataset yields worse performance.


\begin{table}[t]
\centering
\begin{tabular}{lcccc}
\toprule
Method     & F1 & Prec & Rec & Human \\
\midrule
pix2pix \cite{pix2pix2017} & 0.536 & 0.625  & 0.469 & 4.0\% \\
gPb-OWT-UCM \cite{amfm_pami2011} & 0.697 & 0.634 & 0.774 & 4.5\% \\
HED \cite{xie15hed} & 0.782 & 0.779 & 0.785 & 13.0\% \\
{\bf Ours} & {\bf 0.822} & 0.879 & 0.773 & {\bf 19.5}\% \\
\bottomrule
\end{tabular}
\caption{We evaluate contour drawing generation using standard schemes for evaluating boundary detection, modified to allow for larger pixel offsets during matching. Our method outperforms strong baselines for image translation and boundary detection. In the last column, we measure accuracy with an A/B user study. Our generated drawings are able to fool significantly more human subjects. User studies are consistent with our quantitative evaluation, suggesting that boundary detection accuracy is a reasonably proxy for contour drawing generation.}
\label{tab:sketch}
\end{table}

{\bf Perceptual Study}
Besides measuring the F1-score, we also evaluate results with A/B testing in a perceptual study (shown in the last column in Tab~\ref{tab:sketch}). For each algorithm, we present AMT Turkers~\cite{Buhrmester2011AmazonsMT} with a tuple consisting of an image, the generated drawing, and a random human drawing for that image for 5 seconds.  Turkers are then asked to select the one drawn by a human. Past A/B tests for image generation tend to use a shorter presentation time of 1 second~\cite{pix2pix2017,CycleGAN2017}, making it easier to fool a user.

{\bf In-the-Wild Testing}
We also test the generalizability of our method on arbitrary Internet images. Note that the results here are obtained by directly applying the top performing model on the sketch validation set, {\em without any tuning of the hyperparameters}. The qualitative results in Fig \ref{fig:teaser} show that our model has learned general representations for salient contours in the images without content bias and incorporates random perturbations present in human drawings. The generalization power to unseen contents suggests that our method can be applied to other tasks, for instance, salient boundary detection, which is discussed in the following section. Moreover, the generalization to arbitrary contents is also crucial for us to design a human-in-the-loop learning scheme for dataset expansion (Sec \ref{sec:expansion}).

\section{Application to Boundary Detection} \label{sec:boundarydetection}

\begin{figure*}[t]
\begin{center}
\includegraphics[width=1\linewidth]{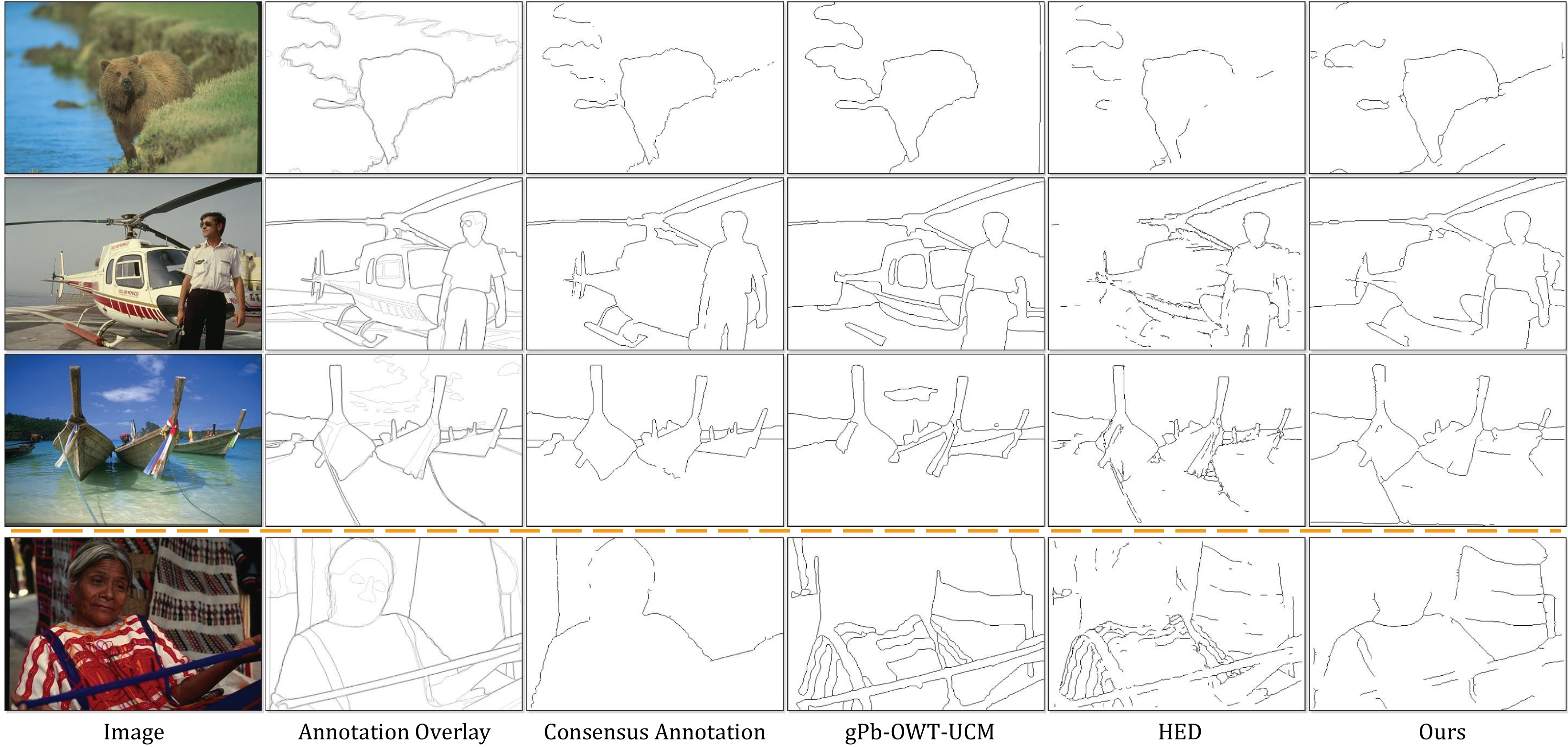}
\end{center}
\vspace{-1em}
\caption{Qualitative results for salient boundary detection. Our method generates complete objects more frequently than competing methods and yet without over generating texture edges. The last row shows a challenging case where all methods fail.}
\label{fig:visualbsds}
\end{figure*}

For an algorithm to generate a contour drawing, it needs first to identify salient edges in an image and this implies that our contour sketch can be re-purposed for salient boundary detection.

{\bf Previous Methods} Edge detectors are precursors to boundary detectors. Those methods \cite{kittler1983accuracy,Canny1986ACA} are usually filter-based and closely related to intensity gradients in images. Since Martin \etal \cite{1273918} first raised the problem of boundary detection, many efforts \cite{amfm_pami2011,Dollr2006SupervisedLO,Ren2008MultiscaleIB,Lim2013SketchTA,Dollr2015FastED} have been devoted to building learning methods upon hand-crafted features. Benchmark performance was improved by a series of deep methods \cite{Shen2015DeepContourAD,Bertasius2015DeepEdgeAM,xie15hed}. However, most of these methods merge the set of annotations to one before training ignoring annotation inconsistency issue. In addition, adversarial training has not yet been applied to boundary detection.


{\bf Salient Boundaries} Image boundaries may be somewhat ambiguous. This can be seen from the open-ended instructions used to annotate BSDS500 \cite{MartinFTM01, amfm_pami2011} (divide the images into 2 to 20 regions),
which often results in inconsistent boundaries labeled by annotators for the same image.
Hou \etal \cite{Hou2013BoundaryDB} studied this inconsistency issue in BSDS500 with a series of experiments. They define {\em orphan} labels to be boundaries labeled by only a single annotator. They then ask human subjects if a particular algorithm's false alarm (predicted boundary pixel that is not matched to any ground-truth) is stronger than a randomly-selected orphan. Subjects selected the false alarm 50.2\% of the time. This seems worrisome as nearly half of the false alarms are stronger than a ground-truth boundary. When repeating this experiment with a consensus boundary pixel rather than an orphan boundary, subjects select the false alarm only 10.9\% of the time. 
Motivated by this observation, Hou \etal suggest evaluating results using only consensus boundary pixels as the ground-truth.

\begin{table}[t]
\centering
\begin{tabular}{lccc}
\toprule
Method     & F1-score & Precision & Recall \\
\midrule

gPb-OWT-UCM \cite{amfm_pami2011} & 0.591 & 0.512   & 0.698 \\
DeepContour \cite{Kokkinos2010BoundaryDU} & 0.615 & 0.540   & 0.714 \\
DeepEdge \cite{Bertasius2015DeepEdgeAM} & 0.651 & 0.608   & 0.700 \\
HED \cite{xie15hed} & 0.665 & 0.580   & 0.778 \\
RCF \cite{Liu2017RicherCF} & 0.693 & 0.621   & 0.784 \\
Ours (w/o pre-training) & 0.637 & 0.627 & 0.646 \\
{\bf Ours (final)} & {\bf 0.707} & 0.706 & 0.708 \\
\bottomrule
\end{tabular}
\caption{Salient (consensus) boundary detection on the BSDS500 dataset, a standard benchmark set up by \cite{Hou2013BoundaryDB}.
}
\label{tab:strongbsds}
\vspace{-1em}
\end{table}

Qualitatively, we find such boundary pixels correspond to salient contours on object boundaries.
This criterion appears to be quite reliable in BSDS, but weak boundaries (of which not all annotators agree) may not be. Many of the weak boundaries may be artifacts caused by original segmentation-like annotation protocols in BSDS. As a result, the standard BSDS benchmark favors algorithms that tend to over-generate boundary predictions so as to have high recall of weak boundaries. Therefore, we focus on salient boundary detection and adopt the evaluation criteria proposed by \cite{Hou2013BoundaryDB}.




{\bf Results} The BSDS500 is also a dataset with 1-to-many mapping. We can directly apply our method to this task. When we train our method on BSDS500, we experiment with two settings: whether we pre-train on our contour drawing dataset. The results are summarized in Tab \ref{tab:strongbsds} and Fig \ref{fig:visualbsds}. When our model is trained only on BSDS500, it performs worse than HED \cite{xie15hed} \& RCF \cite{Liu2017RicherCF}, which are pre-trained on ImageNet. But after fine-tuning, our method notably outperforms HED \& RCF by a sizeble margin. Interestingly, our fine-tuned model learns to generate contours with precise pixel alignment.

\section{Cost-Free Data Expansion} \label{sec:expansion}

The edge alignment in our contour drawing is not as perfect as in BSDS500, but it is this very imperfection that makes data collection much easier, and therefore much more scalable. In this ``deep'' era, both BSDS500 and our current dataset are considered ``small''. Comparing to boundary annotations in BSDS500, our sketch annotations typically contain more details, but it is a much easier and more interesting task than annotating precise boundaries, since we only require loose alignment. During data collection, we frequently received comments like ``this is really fun'', ``I like this {\em game}'' and ``I enjoyed the task''. Motivated by such comments, we further extended the interface to a sketch drawing game with the goal of collecting large scale data for free.



{\bf Prior Work} Von Ahn and Dabbish \cite{Ahn2004LabelingIW} first point out that games can be used to label images. Their game asks a pair of players to guess the label of the same image and extract the common input as the final label. Another game is built by Deng and his colleagues \cite{Deng2013FineGrainedCF} in which they mask an image of an object and ask the player which fine-grained class (e.g. the species of a bird) is present. The player is allowed to erase blobs of the mask at some penalty to better inspect the image. Several other games, WhatsMySketch \cite{Eitz2012HowDH} and QuickDraw \cite{Ha2018ANR}, are built with the idea of letting the player to draw the sketch in order for an artificial intelligence system to recognize it. DrawAFriend \cite{Limpaecher2013RealtimeDA} lets players trace images of celebrities or their friends and send their finished drawings to their friend to guess the identity.

\begin{figure}[t]
\begin{center}
\includegraphics[width=1\linewidth]{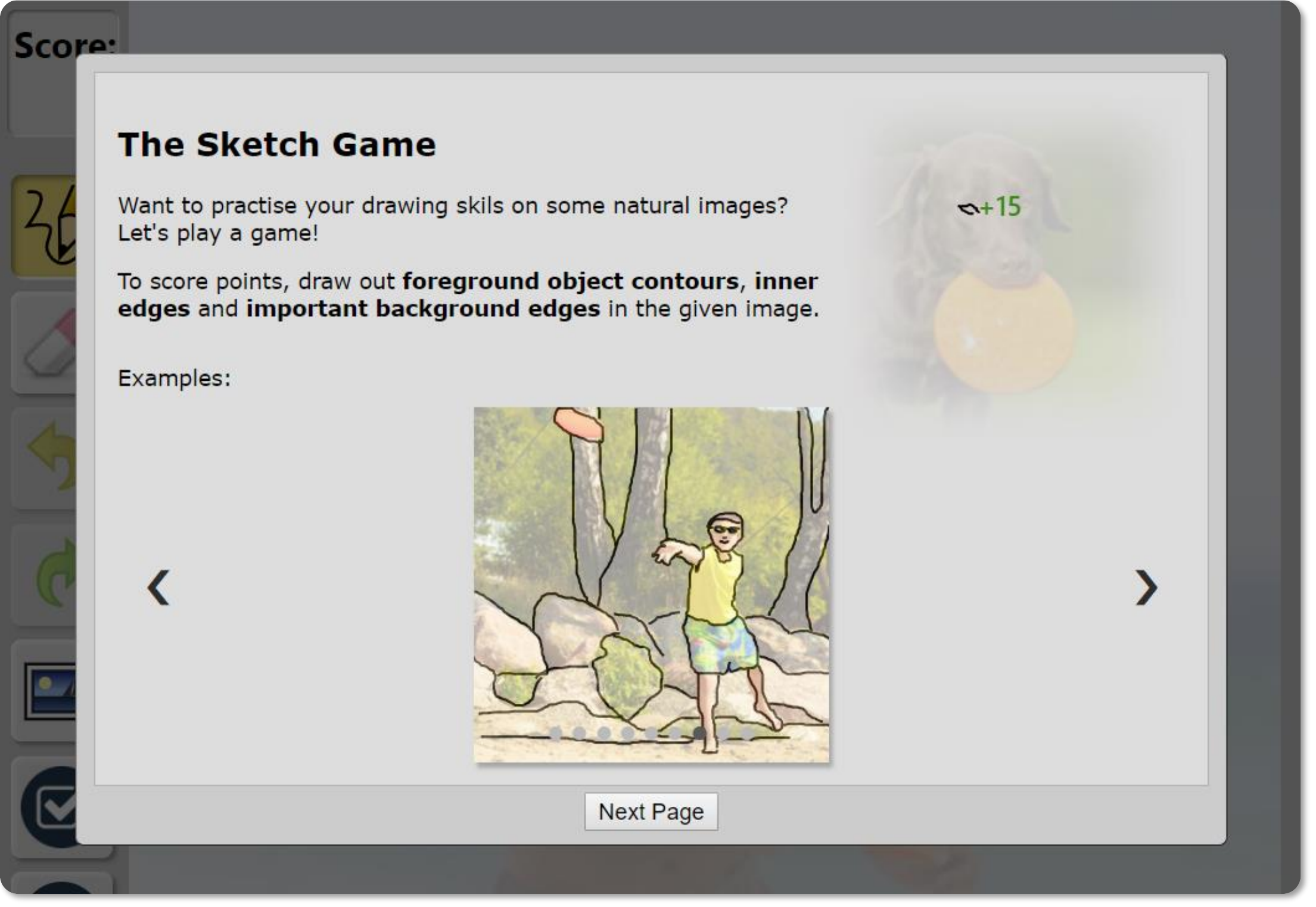}
\end{center}
\vspace{-1em}
\caption{Our sketch game for automatic large scale data collection. The game implements real-time reward/penalty feedback and an automatic quality control mechanism.}
\label{fig:sketchgame}
\end{figure}

{\bf Gaming Interface} The fact that many data-collection games make use of sketches reinforces our observation that drawing itself is an interesting task. Therefore, we develop a game app for scalable data collection (Fig \ref{fig:sketchgame}, demo video can be found on the project website). The challenge here is to set up a game reward system to provide user real-time feedback and also have automatic quality control mechanism for the collected data.
In our game, we first process the image with a boundary detector or a contour generator (as described in Sec \ref{sec:gensketch}), and randomly sample points on the generated boundary map. We then define those points to be reward points, and when a player's drawing matches a reward point, he or she receives the reward associated with the point. We also randomly sample points that are far enough from the generated boundary maps to be penalties points, \ie, when a player stroke is too near a penalty point, he or she loses some of scores. Now the players get instant feedback on how well they draw and the total scores they obtained is a measure of the sketch quality. Note that all reward points are hidden to the players. Since both the reward points and penalty points are sparse, this will not limit the player to precisely follows the algorithm's output. Then we set a cut-off score (as a percentage of the total available rewards) to perform final decision on whether to accept this sketch or not.

{\bf Evaluation} To evaluate this reward system, or {\em AI score} for short, 
we conduct two experiments. For the first experiment, 
we set the sketches in our initial data collection phase (Sec \ref{sec:dataset}) as ground truth, where 5000 of them are manually marked qualified and 1947 of them unqualified. When we use the AI score to rate those sketches, it can identify 90\% of unqualified sketches, showing the capability of rejecting poor submissions. For the second experiment, we collected 100 additional sketches using the game interface, then we manually marked them as qualified or unqualified based on the standard in Sec \ref{sec:dataset}. For this new testing, the AI score identifies 96\% of the unqualified sketches.


In the future, we plan to release the game to the public to build a free sketch collection machine. As the collection process continues, we can keep update our sketch generation models, which allow us to generate more accurate reward points for the game, and make a never-ending sketch collection and generation system by closing the loop.

\section{Conclusion}

In this work, we examine the problem of generating contour drawings from images, introducing a dataset, benchmark criteria, and generation model. From a graphics perspective, this problem generates aesthetically pleasing line drawings. From a vision perspective, contour sketches allow for the exploration of open contours arising from geometric occlusion events. We show that such data can be used to learn low-level representations that can be fine-tuned to produce state-of-the-art results for salient boundary detection. Looking forward, contour sketches appear to be a scalable alternative for collecting geometric visual annotations (potentially through game interfaces).

{\small
\bibliographystyle{ieee}
\bibliography{egbib}
}

\end{document}